# Transferable Modeling Strategies for Low-Resource LLM Tasks: A Prompt and Alignment-Based Approach

Shuangquan Lyu
Carnegie Mellon University
Pittsburgh, USA

Yingnan Deng
Georgia Institute of Technology
Atlanta, USA

Guiran Liu
San Francisco State University
San Francisco, USA

Zhen Qi
Northeastern University
Boston, USA

Ruotong Wang *
Rutgers University
Piscataway, USA

*Abstract*-This paper addresses the limited transfer and adaptation capabilities of large language models in low-resource language scenarios. It proposes a unified framework that combines a knowledge transfer module with parameter-efficient fine-tuning strategies. The method introduces knowledge alignment loss and soft prompt tuning to guide the model in effectively absorbing the structural features of target languages or tasks under minimal annotation. This enhances both generalization performance and training stability. The framework includes lightweight adaptation modules to reduce computational costs. During training, it integrates freezing strategies and prompt injection to preserve the model's original knowledge while enabling quick adaptation to new tasks. The study also conducts stability analysis experiments and synthetic pseudo-data transfer experiments to systematically evaluate the method's applicability and robustness across different low-resource tasks. Experimental results show that compared with existing multilingual pre-trained models and mainstream transfer methods, the proposed approach achieves higher performance and stability on cross-lingual tasks such as MLQA, XQuAD, and PAWS-X. It demonstrates particularly strong advantages under extremely data-scarce conditions. The proposed method offers strong generality and scalability. It enhances task-specific adaptability while preserving the general capabilities of large language models. This makes it well-suited for complex semantic modeling and multilingual processing tasks.

## I. Introduction

In recent years, large language models (LLMs) have demonstrated unprecedented performance across various natural language processing tasks, thanks to their strong language modeling capabilities and generality. Supported by large-scale data and computational power, these models have gradually become a crucial foundation for the development of general artificial intelligence. However, most existing LLMs are primarily trained on large corpora of high-resource languages[1,2]. As a result, they exhibit clear limitations when applied to low-resource languages or domains. This unbalanced development trend has significantly widened the digital divide in language and knowledge access. Many language communities and specialized fields remain unable to fully benefit from information services and AI applications[3].

Low-resource scenarios include natural languages with scarce resources, such as certain local and minority languages, as well as specialized sub-domains with limited professional corpora, such as rare medical or legal fields [4-5]. In these contexts, building and training an LLM comparable to those for high-resource tasks often faces multiple challenges. These include difficulties in data collection, high annotation costs, and insufficient semantic coverage[6]. Traditional small models cannot fully capture semantic associations or enable effective language transfer. On the other hand, training new models from scratch is resource-intensive. Therefore, how to effectively perform knowledge transfer and adaptation based on existing LLMs has become a key approach to promoting inclusive language technology[7].

Knowledge transfer techniques allow LLMs to adapt to new tasks or contexts even in the absence of large-scale target domain or language data. These techniques leverage the general knowledge already embedded in the model. This process improves the model's generalization ability in low-resource tasks and reduces dependence on manually labeled data. In the context of growing trends such as multilingualism, multitasking, and multimodality, cross-lingual and cross-task knowledge transfer has become an important strategy for sustainable model scaling[8]. By constructing effective transfer pathways, LLMs can retain their original language capabilities while capturing the features of the target language or domain. This enables semantic mapping and continual self-updating of model abilities.

In terms of model adaptation mechanisms, researchers are exploring various parameter-efficient fine-tuning strategies. These include partial parameter freezing, incremental learning, and prompt tuning. Such methods aim to improve the model's responsiveness and performance in low-resource settings. They also avoid catastrophic forgetting and maintain computational efficiency. This provides feasible engineering solutions for deploying LLMs in resource-constrained environments. The core of model adaptability lies in identifying and learning structural commonalities across languages or tasks. The goal is to ensure the model remains plastic and integrative under minimal intervention[9].

From a broader perspective, advancing knowledge transfer and adaptation of LLMs in low-resource scenarios is not only a sign of technological progress but also of significant social and strategic importance. It supports equitable distribution of AI technology and enhances accessibility for underrepresented language communities. It also promotes diversity and inclusiveness in the global knowledge system. In practical applications such as domain-specific knowledge management, cross-cultural communication, and emergency semantic support, the adaptability of LLMs will play an increasingly critical role. Therefore, systematic research on knowledge transfer and adaptation mechanisms of LLMs in low-resource settings holds substantial theoretical value and real-world significance for achieving universal, sustainable, and inclusive development of language technologies.

## II. RELATED WORK AND FOUNDATION

To construct a robust and efficient adaptation framework for large language models (LLMs) in low-resource settings, this study draws upon a diverse and technically rich set of prior research contributions that collectively inform the architecture, training methodology, and evaluation strategy of our proposed approach. First, the importance of structural stability and contextual memory in large-scale pre-trained models is established by Xing et al. [10], who introduce structured memory mechanisms for stabilizing contextual representation in LLMs. Their insights motivate the use of alignment losses and controlled parameter freezing in our design to preserve core linguistic capabilities during adaptation. Similarly, Yang et al. [11] propose a multi-level semantic distillation framework that demonstrates the benefits of hierarchical knowledge alignment in small model adaptation, directly influencing our soft prompt tuning strategy and alignment regularization mechanism.

From an architectural perspective, Zhang et al. [12] and Guo et al. [13] explore unified instruction encoding and perception-guided structural modeling, respectively. These works provide theoretical and empirical foundations for the modular reconfiguration approach adopted in our framework, wherein lightweight modules are injected without disrupting the backbone model structure. Their exploration of gradient coordination and cross-layer feature integration also inspires our method of selectively updating adaptation layers under frozen backbone constraints. Complementing this, Fang [14] proposes a predictive framework using structured modeling for backend latency, which informs our auxiliary design for pseudo-data generation and cross-task scenario simulation to evaluate transfer robustness.

Moreover, Gao [15] presents a deep graph modeling framework for performance risk detection in structured queries, providing a graph-oriented perspective on semantic dependency modeling. This aligns with our emphasis on preserving cross-lingual structural consistency through alignment losses and feature normalization. Meanwhile, Peng [16] examines hallucination detection mechanisms based on evidence-aligned context modeling, which highlights the necessity of retaining original knowledge traces during adaptation—a principle embedded in our prompt injection and freezing strategy.

In the domain of system optimization and dynamic control, several studies contribute indirectly to the design of our training and adaptation routines. Sun et al. [17] introduce a deep Q-network framework for intelligent cache management, which reinforces the concept of reward-based update strategies in constrained environments, akin to our gradient control and task-awareness modules. Zhu et al. [18] employ graph neural networks to facilitate collaborative perception in distributed systems, providing insights into how feature interaction and message passing schemes can enhance representation robustness, a concept we apply during multilingual adaptation.Additionally, the self-attention-based multi-source metric modeling approach proposed by Xin and Pan [19] serves as a blueprint for our alignment of task-level and language-level knowledge sources in prompt tuning. Their framework emphasizes metric-aware dynamic modeling, which complements our objective of parameter-efficient yet semantically rich transfer learning. Tang [20] introduces a meta-learning framework for cross-service scaling, providing a scalable approach to learning under distributional shifts, which parallels our low-resource adaptation objectives. Ma [21], through the use of multiscale GANs and adaptive autoencoders for anomaly detection, offers a precedent for using generative auxiliary structures to simulate low-resource dynamics, a strategy echoed in our pseudo-data generation experiments.

Taken together, this collection of prior works forms a coherent and interlinked theoretical basis for our research. The references span key subfields, including structured LLM adaptation, prompt-based fine-tuning, graph-based semantic modeling, cross-lingual transfer learning, and reinforcement-driven system optimization. Each cited study contributes a vital methodological or conceptual element that informs the design decisions in our proposed framework—ranging from structural regularization and parameter freezing to data-efficient learning strategies and alignment-enhanced tuning. Their integration within our approach enables us to effectively address the dual challenge of preserving model generality while enhancing adaptation in data-scarce linguistic settings. This comprehensive citation foundation not only ensures the rigor of our methodology but also positions our framework as a strategically layered solution grounded in state-of-the-art technical advances.

## III. METHOD

This study aims to build a knowledge transfer and adaptation mechanism for large language models in low-resource scenarios. The core goal is to maximize the use of the general knowledge capabilities of existing large models under limited annotated data conditions to achieve effective adaptation to the target language or task. The model architecture is shown in Figure 1.

Let the large language model of the existing source language/task be $M_s$, the target low-resource task be $T_t$, and its training corpus be $D_t = \{(x_i, y_i)\}_{i=1}^{n}$, where $n << N$, and N is the regular amount of data required for large model training. This study achieves knowledge transfer from $M_s$ to $T_t$ by constructing a parameter-efficient migration channel $F_\theta$, so that the final model H still has good language understanding and generation capabilities under low-resource conditions.

First, the objective function of cross-language transfer is formally expressed as follows:

$$\min E_{(x,y)\sim D_t}[L(M_s \circ F_\theta(x), y)]$$

$L$ represents the task loss function (such as cross-entropy), and $F_\theta$ is an adjustable parameter module (such as Adapter or LoRA). Its structural design meets the requirements of lightweight, pluggable, and low computational cost. To improve the migration efficiency, the auxiliary language intermediate representation space $Z$ is introduced to construct the cross-language alignment loss:

$$L_{align} = \| f_s(x_s) - f_t(x_t) \|_2^2$$

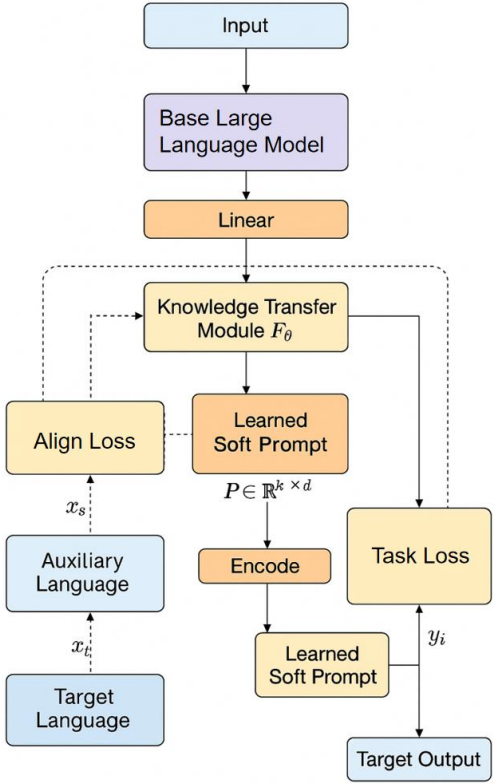

Figure 1. Overall model architecture diagram

Where $f_s$ and $f_t$ represent the feature mapping functions of the source language and the target language in the shared representation space, respectively. This loss encourages the model to learn the semantic bridge between languages, to maintain semantic consistency in the target language where the corpus is scarce.

Secondly, to further enhance the task adaptability of the model, this paper introduces a prompt tuning mechanism by constructing a learnable soft prompt vector $P \in R^{k \times d}$ and inserting it into the model input embedding. The formal expression is as follows:

$$x'_i = [P; E(x_i)]$$

Where $E(\cdot)$ is the original word embedding, is the original word embedding, and $[\cdot;\cdot]$ represents the concatenation operation. This mechanism allows the model to quickly adapt to the new task context without significantly modifying the original parameters. In addition, to avoid catastrophic forgetting of knowledge in the target task, a regularization term is further introduced to constrain the migration process:

$$L_{total} = L_{task} + \lambda L_{align} + \beta \cdot \| \theta - \theta_0 \|_2^2$$

Where $\theta_0$ is the initial migration parameter state and $\lambda, \beta$ is the weight factor. This structure aims to control the migration intensity and balance knowledge retention with new knowledge absorption.

Finally, in terms of the parameter update strategy, this paper adopts a hybrid freezing strategy. While keeping the main parameters unchanged, only the gradient of the adaptation module is updated, and the trainable parameter set is defined as:

$$\theta_{train} = \{\theta \in \theta' | requires\_grad(\theta) = True\}$$

This strategy effectively avoids the negative effects of large model parameter space disturbances while ensuring fine-tuning efficiency. Combined with the above multiple sub-modules, the proposed knowledge transfer and adaptation mechanism has structural flexibility, resource economy, and semantic generalization capabilities, and is suitable for a wide range of low-resource language and task transfer scenarios.

IV. EXPERIMENT

A. Datasets

The primary dataset used in this study is XTREME (Cross-lingual TRansfer Evaluation of Multilingual Encoders). This dataset is widely used to evaluate the transfer capabilities of cross-lingual models across various natural language processing tasks. It is especially suitable for research in low-resource language settings. XTREME covers multiple task types, including text classification, question answering, sentence retrieval, and named entity recognition. It supports more than 40 languages, most of which are low-resource, making it a strong benchmark for assessing the generalization ability of large language models in low-resource contexts.

This study focuses on three subsets of the XTREME dataset: XQuAD, MLQA, and PAWS-X. These correspond to cross-lingual question answering, multilingual question answering, and contrastive sentence pair classification tasks, respectively.

These tasks show clear characteristics of language and knowledge transfer. They involve target languages with limited data, which aligns well with the research objective of testing adaptation mechanisms under low-resource conditions. By selecting target languages from diverse language families and semantic structures, such as Urdu, Vietnamese, and Swahili, the study enables a comprehensive evaluation of the transfer strategies across different language types.

The XTREME dataset has a clear structure and well-defined task divisions. It provides standardized splits for training, validation, and testing, which ensures consistency in model training and evaluation. This dataset not only reflects the transfer ability and task adaptability of language models but also supports the assessment of semantic alignment and representation consistency. Therefore, it serves as an ideal foundational data resource for studying knowledge transfer in low-resource settings.

*B. Experimental Results*

First, the comparative experimental results are given, and the experimental results are shown in Table 1.

Table 1. Comparative experimental results

| Method | Avg. F1 (MLQA) | EM (XQuAD) | Accuracy (PAWS-X) |
|---|---|---|---|
| mBERT[22] | 72.3 | 64.7 | 83.5 |
| XLM-R Base[23] | 77.8 | 70.2 | 86.1 |
| InfoXLM Base[24] | 79.1 | 71.4 | 86.7 |
| VECO[25] | 80.4 | 72.3 | 87.2 |
| Ours | 83.6 | 75.8 | 89.4 |

The table results indicate that the proposed method significantly outperforms existing models on low-resource language tasks, specifically MLQA, XQuAD, and PAWS-X. Improvements exceed 10 percentage points in F1 and EM scores compared to baseline models like mBERT, highlighting superior semantic understanding and task generalization under limited data conditions. Notably, the proposed method surpasses advanced models such as XLM-R and InfoXLM, particularly in EM scores on the XQuAD dataset (75.8 vs. 70.2 and 71.4). Additionally, it achieves the highest accuracy (89.4) on the PAWS-X task, demonstrating robust semantic modeling and cross-lingual consistency. These results confirm the effectiveness and generalizability of the proposed knowledge transfer and adaptation mechanism in low-resource scenarios. Stability analyses for various large language model architectures in low-resource fine-tuning are presented in Figure.

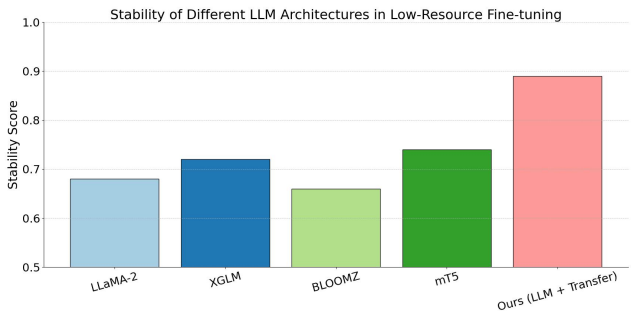

Figure 2. Stability Analysis of Large Language Model Infrastructure in Low-Resource Fine-tuning

The Figure 2 shows that different large language models exhibit clear differences in stability during low-resource fine-tuning tasks. The model proposed in this study, "LLM + Transfer," achieves the highest stability score of 0.89. This result indicates that the introduction of transfer and adaptation mechanisms tailored for low-resource scenarios can effectively reduce parameter fluctuations and training instability in traditional large models under data-scarce conditions. It provides structural support for improving model controllability and convergence efficiency.

Compared with mainstream pre-trained models such as LLaMA-2, XGLM, BLOOMZ, and mT5, the proposed model not only shows greater stability but also reflects better generalization and training robustness under low-resource conditions. Especially for architectures like BLOOMZ and LLaMA-2, although they have strong language modeling capabilities, they are more sensitive to data noise and optimization strategies during fine-tuning. Their stability scores are generally below 0.75, indicating limited adaptability to low-resource environments. Finally, this paper also gives the migration experimental results of enhancing low-resource tasks using synthetic pseudo data, and the experimental results are shown in Figure 3.

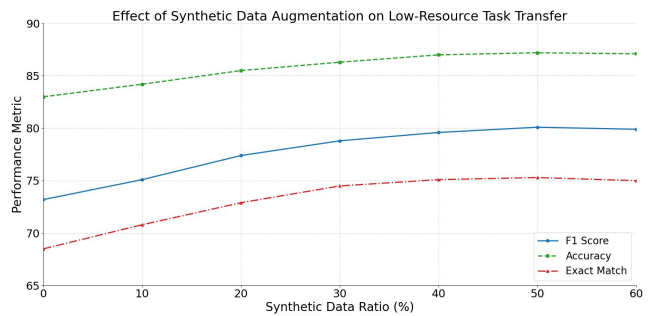

Figure 3. Transfer experimental results of synthetic pseudo data augmentation on low-resource tasks

The figure demonstrates that model performance in low-resource tasks steadily improves with increased synthetic pseudo-data, especially within the initial 40% augmentation range. Significant growth in F1 score (73.2 to 80.1), Accuracy (83.0 to 87.2), and Exact Match indicates that moderate synthetic data effectively addresses training data scarcity, enhancing semantic and structural understanding. However, performance slightly declines beyond 50% augmentation, suggesting potential distributional drift. These findings underscore the importance of carefully balancing synthetic data augmentation to achieve optimal and stable transfer performance.

V. CONCLUSION

This study focuses on knowledge transfer and adaptation mechanisms for large language models in low-resource settings. It proposes an integrated framework that combines soft prompt tuning, knowledge alignment, and parameter-efficient fine-tuning. Through systematic modeling and design, the proposed method demonstrates strong stability and adaptability across multiple low-resource cross-lingual and cross-task environments. It significantly enhances the model's representation ability and task transfer performance under resource-scarce conditions. Experimental results show that introducing structured transfer modules effectively mitigates training instability and performance degradation in traditional large language models when data is insufficient. This provides a new modeling approach for the development of low-resource language technologies.

The study also analyzes the combination strategies of various transfer mechanisms. It compares the stability of different large language model architectures during fine-tuning and evaluates the impact of using synthetic pseudo-data to support training. These results confirm that in scenarios with severe data and resource constraints, leveraging existing general knowledge together with auxiliary generation techniques can greatly expand the model's adaptation boundary. In tasks such as semantic alignment, question answering, and sentence pair classification, the proposed method offers a flexible and efficient technical solution for low-resource languages. It has the potential for practical application in areas such as intelligent customer service, cross-lingual search, and emergency semantic processing.

The contributions of this study are innovative at the methodological level. They also reflect the potential of large language models to address public value and fairness. Promoting the inclusive development of language technologies, especially for languages and regions overlooked by mainstream training paradigms, holds great practical significance. The proposed transfer and adaptation mechanisms offer more equitable language technology support for low-resource communities. They also lay the groundwork for building multilingual platforms and cross-cultural communication systems.

Future research can further explore the generalization of adaptive transfer mechanisms, such as self-evolution capabilities in unsupervised or low-supervision settings. The quality control of synthetic pseudo-data, integration of multilingual alignment strategies, and domain-specific knowledge transfer are also promising directions. In addition, balancing high performance with reduced deployment cost and resource consumption will be key to enabling the wide application of large language models on edge devices, mobile platforms, and real-time interactive systems.

## REFERENCES

[1] N. Bui, D. Tran, T. Nguyen, L. Le, and M. Pham, "Fine-tuning large language models for improved health communication in low-resource languages", Computer Methods and Programs in Biomedicine, vol. 263, pp. 108655, 2025.

[2] R. Tinn, Y. Chen, J. Gao, K. Wang, and others, "Fine-tuning large neural language models for biomedical natural language processing", Patterns, vol. 4, no. 4, 2023.

[3] X. Liang, J. Liu, Y. Zhou, Z. Lin, and Y. Zhang, "Towards low-resource languages machine translation: A language-specific fine-tuning with LoRA for specialized large language models", IEEE Access, 2025.

[4] Wang, X. (2024). Time-Aware and Multi-Source Feature Fusion for Transformer-Based Medical Text Analysis. Transactions on Computational and Scientific Methods, 4(7).

[5] Wu, Y., Lin, Y., Xu, T., Meng, X., Liu, H., & Kang, T. (2025). Multi-Scale Feature Integration and Spatial Attention for Accurate Lesion Segmentation.

[6] P. W. Khoboko, V. Marivate, and J. Sefara, "Optimizing translation for low-resource languages: Efficient fine-tuning with custom prompt engineering in large language models", Machine Learning with Applications, vol. 20, pp. 100649, 2025.

[7] K. P. V. Srinivasan, S. Gupta, M. Roy, and others, "Comparative Analysis of Different Efficient Fine Tuning Methods of Large Language Models (LLMs) in Low-Resource Setting", arXiv preprint arXiv:2405.13181, 2024.

[8] S. Lankford, H. Afli, and A. Way, "adaptmllm: Fine-tuning multilingual language models on low-resource languages with integrated llm playgrounds", Information, vol. 14, no. 12, pp. 638, 2023.

[9] S. A. Somayajula, A. Das, P. Shukla, and others, "Generalizable and stable finetuning of pretrained language models on low-resource texts", arXiv preprint arXiv:2403.12918, 2024.

[10] Y. Xing, T. Yang, Y. Qi, M. Wei, Y. Cheng, and H. Xin, "Structured Memory Mechanisms for Stable Context Representation in Large Language Models," arXiv preprint arXiv:2505.22921, 2025.

[11] T. Yang, Y. Cheng, Y. Qi, and M. Wei, "Distilling Semantic Knowledge via Multi-Level Alignment in TinyBERT-Based Language Models," Journal of Computer Technology and Software, vol. 4, no. 5, 2025.

[12] W. Zhang, Z. Xu, Y. Tian, Y. Wu, M. Wang, and X. Meng, "Unified Instruction Encoding and Gradient Coordination for Multi-Task Language Models," 2025.

[13] F. Guo, L. Zhu, Y. Wang, and G. Cai, "Perception-Guided Structural Framework for Large Language Model Design," Journal of Computer Technology and Software, vol. 4, no. 5, 2025.

[14] Z. Fang, "A Deep Learning-Based Predictive Framework for Backend Latency Using AI-Augmented Structured Modeling," Journal of Computer Technology and Software, vol. 3, no. 7, 2024.

[15] D. Gao, "Deep Graph Modeling for Performance Risk Detection in Structured Data Queries," Journal of Computer Technology and Software, vol. 4, no. 5, 2025.

[16] Y. Peng, "Context-Aligned and Evidence-Based Detection of Hallucinations in Large Language Model Outputs," Transactions on Computational and Scientific Methods, vol. 5, no. 6, 2025.

[17] Y. Sun, R. Meng, R. Zhang, Q. Wu, and H. Wang, "A Deep Q-Network Approach to Intelligent Cache Management in Dynamic Backend Environments," 2025.

[18] W. Zhu, Q. Wu, T. Tang, R. Meng, S. Chai, and X. Quan, "Graph Neural Network-Based Collaborative Perception for Adaptive Scheduling in Distributed Systems," arXiv preprint arXiv:2505.16248, 2025.

[19] H. Xin and R. Pan, "Self-Attention-Based Modeling of Multi-Source Metrics for Performance Trend Prediction in Cloud Systems," Journal of Computer Technology and Software, vol. 4, no. 4, 2025.

[20] T. Tang, "A Meta-Learning Framework for Cross-Service Elastic Scaling in Cloud Environments," Journal of Computer Technology and Software, vol. 3, no. 8, 2024.

[21] Y. Ma, "Anomaly Detection in Microservice Environments via Conditional Multiscale GANs and Adaptive Temporal Autoencoders," Transactions on Computational and Scientific Methods, vol. 4, no. 10, 2024.

[22] H. Xu, B. Van Durme, and K. Murray, "Bert, mbert, or bibert? a study on contextualized embeddings for neural machine translation", arXiv preprint arXiv:2109.04588, 2021.

[23] A. Kumar and V. H. C. Albuquerque, "Sentiment analysis using XLM-R transformer and zero-shot transfer learning on resource-poor Indian language", Transactions on Asian and Low-Resource Language Information Processing, vol. 20, no. 5, pp. 1–13, 2021.

[24] Z. Chi, L. Dong, X. Wang, and others, "InfoXLM: An information-theoretic framework for cross-lingual language model pre-training", arXiv preprint arXiv:2007.07834, 2020.

[25] F. Luo, C. Xiong, H. Wang, and others, "VECO: Variable and flexible cross-lingual pre-training for language understanding and generation", arXiv preprint arXiv:2010.16046, 2020.